\documentclass[runningheads]{llncs}
\usepackage[T1]{fontenc}
\usepackage{graphicx}
\usepackage{url}
\usepackage{booktabs}
\usepackage{hyperref}
\usepackage{color}

\hypersetup{
    pdfinfo={
        Title={A Theoretically Grounded Benchmark for Semantic Evaluation of Machine Common Sense},
        Author={Henrique Santos, Ke Shen, Alice M. Mulvehill, Yasaman Razeghi, Deborah L. McGuinness, Mayank Kejriwal},
        Keywords={commonsense reasoning, commonsense semantics, language representation models, question answering, gordon-hobbs theory}
    }
}

\begin{document}
\title{A Theoretically Grounded Benchmark for Semantic Evaluation of Machine Common Sense}
\titlerunning{Theoretically-Grounded Commonsense Reasoning}
%
\author{Henrique Santos\inst{1} \and
Ke Shen\inst{2} \and
Alice M. Mulvehill\inst{1} \and
Yasaman Razeghi\inst{3} \and\\
Deborah L. McGuinness\inst{1} \and
Mayank Kejriwal\inst{2}}
\authorrunning{H. Santos et al.}
%
\institute{Rensselaer Polytechnic Institute, Troy NY, USA \and
University of Southern California, Los Angeles CA, USA \and
University of California, Irvine CA, USA}
\maketitle              
\begin{abstract}
    Achieving machine commonsense has been a longstanding problem within Artificial Intelligence. Thus far, benchmarks that are grounded in a theory of commonsense, and can be used to conduct rigorous, \emph{semantic} evaluations of commonsense reasoning (CSR) systems have been lacking.  We propose the first such benchmark, called \emph{Theoretically-Grounded Commonsense Reasoning (TG-CSR)}. TG-CSR is modeled as a set of question-answering instances, with each instance grounded in a semantic category of commonsense, such as space, time, and emotions. The benchmark is \emph{few-shot} i.e., only a few training and validation examples are provided in the public release to preempt overfitting problems. Evaluations suggest that, due to its semantic rigor, the benchmark is challenging even for billion-parameter statistical models that have achieved near-human performance on other datasets not explicitly designed using commonsense semantics.

    {\bf Public Release:} \url{https://codalab.lisn.upsaclay.fr/competitions/3080}
    
    {\bf Website:} \url{https://usc-isi-i2.github.io/TGCSR/}
    
\end{abstract}
\section{Introduction}

\begin{figure}
\centering
\includegraphics[width=\linewidth]{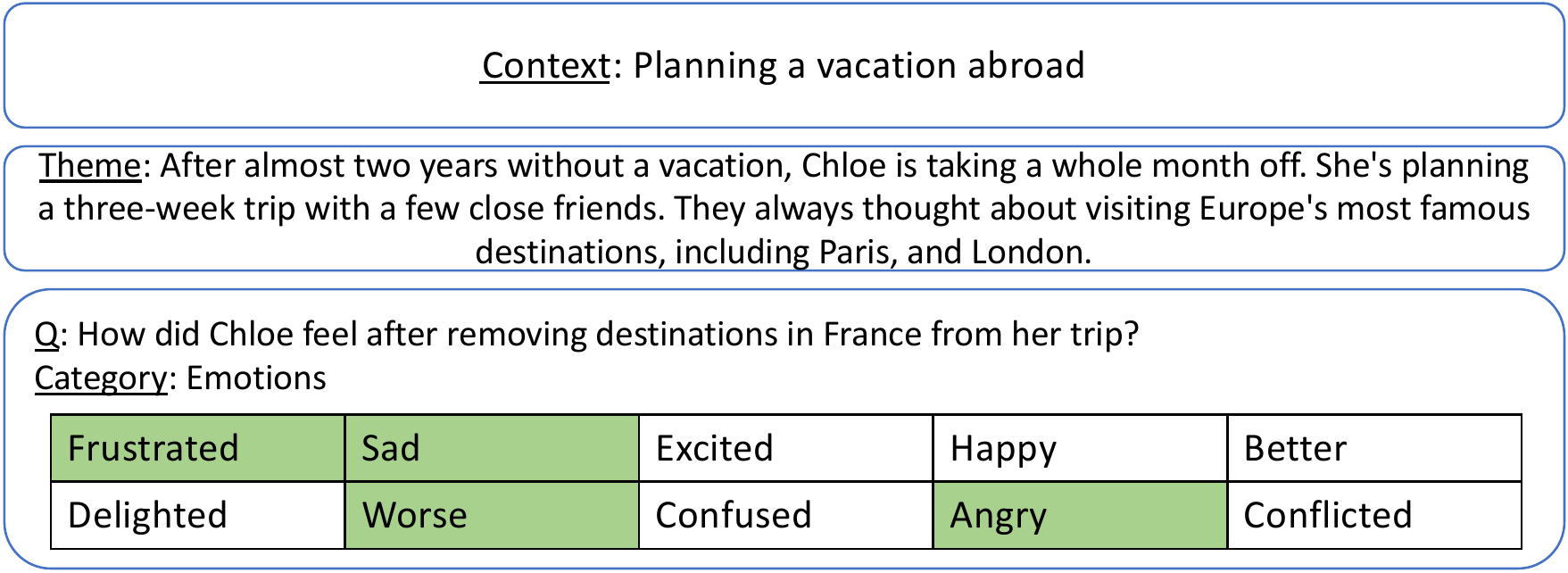}
\caption{An example question (designed for the ``emotions'' category in the Gordon-Hobbs theory) using a ``vacationing abroad'' context in TG-CSR.
A model would be expected to provide a binary answer (e.g., Yes/No) to indicate whether a candidate answer is a suitable response to the question. Ground-truth ``Yes'' answers for this question are shaded in green. More details on benchmark construction and preliminary evaluation are provided in Sections \ref{sec:bcon} and \ref{sec:beval}, respectively.}
\label{fig:overview}
\end{figure} 

Developing machines with commonsense reasoning (CSR) abilities is a longstanding challenge in the Artificial Intelligence community~\cite{davis2015commonsense,Cyc}. Current CSR benchmarks largely use multiple-choice question-answering (QA) instances to evaluate machine commonsense~\cite{santos_exploring_2021}. Unfortunately, many of these QA benchmarks have been constructed in an ad hoc fashion, with little evidence that they are grounded in a formal theory of commonsense, such as the influential theory proposed by Gordon and Hobbs~\cite{gordon2017}. Because the benchmarks are not theoretically grounded, they cannot be used to conduct a \emph{semantic} evaluation of machine learning models on CSR. Additionally, there is some evidence that the state-of-the-art CSR models (which are usually transformer neural networks with billions of parameters) may be `fitting' to a specific QA benchmark's training partition by picking up on subtle and statistical, but semantically irrelevant, features, to achieve good performance on the testing partition~\cite{kejriwal2020fine,kejriwal2022designing}.


In this paper, we propose a benchmark called \emph{Theoretically-Grounded Commonsense Reasoning (TG-CSR)} that is intended to systematically evaluate machine commonsense in a few-shot QA setting. We refer to TG-CSR as \emph{theoretically grounded} because, as discussed below, questions in the benchmark are always designed with respect to a sub-set of categories that were originally proposed and refined by Gordon and Hobbs~\cite{gordon2017}. Importantly, these categories provide semantics for the QA instances in TG-CSR, as discussed subsequently in Section \ref{sec:MCSOnto}. 
The full benchmark is split into four datasets, each of which is associated with a unique theme and context. The context is a short phrase that provides a broad topical interpretation or context for the questions. An example used in the current benchmark release, illustrated in Figure \ref{fig:overview}, is “planning a vacation abroad”. In a similar vein, the theme provides background details, including constraints or motivations of a more specific context. It can be understood as an instantiation of the context for a specific setting. 

Questions and candidate answers for each question are always related to the theme and grounded in one of nine possible fundamental commonsense categories selected from the Gordon-Hobbs theory\footnote{These include ``Time'', ``Space'', ``Scheduling'', ``World States'', ``Physical Entities'', ``Activities'', ``Goals'', ``Values and Quantities'', and ``Emotions''.}.  These categories provide a formal semantics for the benchmark, making TG-CSR one of the first and only benchmarks, to the best of our knowledge, that seeks to evaluate CSR systems in a theoretically grounded manner that includes semantics as a  first-class citizen.




Additionally, unlike existing CSR benchmarks, which are released in a single-phase and prone to overfitting by language models shortly after release, TG-CSR proactively preempts such overfitting. We do this by releasing the benchmark in four \emph{progressively more difficult} phases to ensure that systems are not relying merely on statistical principles but on fundamental semantic aspects of CSR. Currently, we have released the first of the four contexts and themes (specifically, the ``vacationing abroad'' context), split into training, development, and test partitions. Labels are provided for the training and development partitions but withheld for the test partition, which also allows us to mitigate against observer bias. Later phases, all slated for release within 2022 at regular intervals, will use different contexts and themes, with the training and development sets getting relatively smaller in each phase. The final phase will be \emph{zero shot}, with no training and development sets provided. Finally, TG-CSR is hosted on a public and free leaderboard, making it discoverable and sustainable for the future. 


\section{Background and Related Work}

Multiple-choice QA has emerged as a \textit{de facto} standard for evaluating machine commonsense~\cite{mitra2020}. While there were likely several reasons for this occurrence, we identify three main ones: the first-mover effect of early organized efforts~\cite{bentivogli2011}, ease of conducting automated evaluations using multiple-choice QA benchmarks, and funder-driven effects (e.g., the DARPA Machine Common Sense program~\cite{darpa2018} primarily used multiple-choice QA to evaluate the progress of funded teams on adult commonsense reasoning).

Recently, however, specific multiple-choice CSR benchmarks have come under criticism~\cite{elazar-etal-2021-back} and researchers have used adversarial attack techniques (among others) to show that gauging the true machine commonsense ability of these language representation models is currently problematic~\cite{wang2019}. Another major issue with many current benchmarks is that they are often broadly constructed, with only loose or even ad hoc grounding in rigorous commonsense theories, such as those inspired by cognitive science. In contrast, our benchmark takes a semantic, rather than ad hoc approach, by using ontological categories (such as time and space) to guide the construction and evaluation of machine commonsense.

Although some benchmarks exist that are specifically designed to test a model’s capabilities in number estimation, causal reasoning, social reasoning, and more, their semantics are not clearly defined, and their content can be overly diffuse. For example, the Social IQa benchmark purports to test social interactions~\cite{sap2019a}, but arguably, covers a broad range of human abilities and commonsense sub-categories that are not clearly defined. Either way, without a clear semantic framework providing a formal basis for the content of the benchmark, one can only guess from the description (or manually determine the semantics by reading all the thousands of questions, which is obviously infeasible and non-scalable). Similarly, ATOMIC~\cite{sap_atomic_2019} is a dataset for evaluating CSR capabilities on questions involving if-then reasoning, using nine proposed relation types. However, it is not clear if these relations are representative in terms of commonsense reasoning coverage. Cosmos QA~\cite{huang_cosmos_2019} is a benchmark for evaluating text comprehension that requires contextual commonsense reasoning. Questions and answers were collected from four broad categories, including ``causes'', ``effects'', and ``facts'', which are claimed to cover the nine categories in \cite{sap_atomic_2019}. However, there is no evidence of the rigor of how these categories were used or the formal semantics of the categories. 

An example of a semantically grounded commonsense resource is CycIC\footnote{\url{https://leaderboard.allenai.org/cycic/submissions/about}}, which is a benchmark derived from the Cyc platform~\cite{Cyc}, containing sentences to be evaluated as true/false. While the dataset is annotated using some selected categories, including ``norms'', ``theory of mind'', and more, the semantics of these categories are under-defined and more important, the role of each category in the creation of the sentences is not specified. 

In general, the lack of rigorous construction methods (see Paullada et al.~\cite{paullada2021} for a review of methods used to create machine learning datasets), and loose theoretical grounding of many existing benchmarks makes it difficult to make the case that a machine commonsense system is either complete (at least, to the extent that is understood in studies of human cognition), or that it has a grasp of the foundational aspects (such as the semantics) of commonsense tasks and questions.

While the problem of how humans think about things in the world has been extensively studied~\cite{hayes1978naive,allenGeneralTheoryAction1984,hobbs_formal_1985,davis_representations_1990}, there is considerable interest in both AI and cognitive science to better categorize the different kinds of CSR that humans often rely on to navigate everyday life. Brachman and Levesque~\cite{brachman_toward_2022} have argued that CSR, so far, has been mostly studied in support of narrow problems, being unable to consistently achieve human performance when dealing with open-world semantics. For them, one of the core open research questions is what ontological frameworks are critical to build into an AI system. Research efforts reported in recent literature indicate that considerable progress on the axiomatization of commonsense reasoning has been made. Specifically, Gordon and Hobbs performed a comprehensive study of representational requirements for strategic planning. Planning strategies were collected and formally represented in ten different domains. They found that 988 terms could be considered common in all analyzed domains, further clustered them into 48 representational areas, and concluded that eight areas are examples of foundational categories that are involved in human commonsense reasoning and are also commonly used in knowledge representation research~\cite{gordon2017}. However, what has been lacking thus far is a  benchmark that uses these categories to evaluate the performance of advanced CSR systems (such as transformer-based language models) in a theoretically grounded, semantics-first manner. TG-CSR was designed and released with precisely this purpose in mind. 



\section{Building a Machine-Interpretable Taxonomy for Commonsense Reasoning}\label{sec:MCSOnto}

One challenge preventing the sound evaluation of machine commonsense reasoning is the lack of a formal unified representation of the various types of reasoning associated with human commonsense thinking. In contrast, TG-CSR is designed by grounding its questions and answers in the nine categories derived from the 48 representational areas defined in the Gordon-Hobbs theory. The first four are from the set of foundational areas: ``time'', ``space'', ``objects'' (we call these ``physical entities''), and ``values and quantities''. They were selected based on a recent experiment with human annotators that resulted in higher agreement when using these representational areas as labels~\cite{santosExperimentalStudyMeasuring2021}. The next four areas: ``activities'', ``goals'', ``scheduling'' (all intuitively related to ``events''), and ``states'' (we call these ``world states'') were also selected because of high agreement in the same experiment. The ninth and final representational area included in our research is ``emotions'', which was included as a way of evaluating at least one psycho-social aspect of CSR due to its importance.

As previously mentioned, some of these foundational areas have been extensively explored in knowledge representation efforts, including in the Semantic Web~\cite{Little:20:TOO}. However, to the best of our knowledge,  these efforts have not been leveraged collectively to develop a joint machine-interpretable representation that can serve as a formalization of human commonsense reasoning.

This section provides an overview of an initial representation of commonsense reasoning as an ontology. We start from the Gordon-Hobbs theory, focusing on the nine representational areas and their associated terms. While we don't claim completeness, we review and consider the incorporation of certain existing taxonomies, vocabularies, and ontologies as an envisioned unified representation of common sense. For the remainder of this article, we refer to ``representational areas'' as ``categories'', and to ``terms'' as ``concepts'', when used in the context of our knowledge representation and data annotation efforts.

\subsection{Representational areas}

\begin{figure}
\centering
\includegraphics[width=\linewidth]{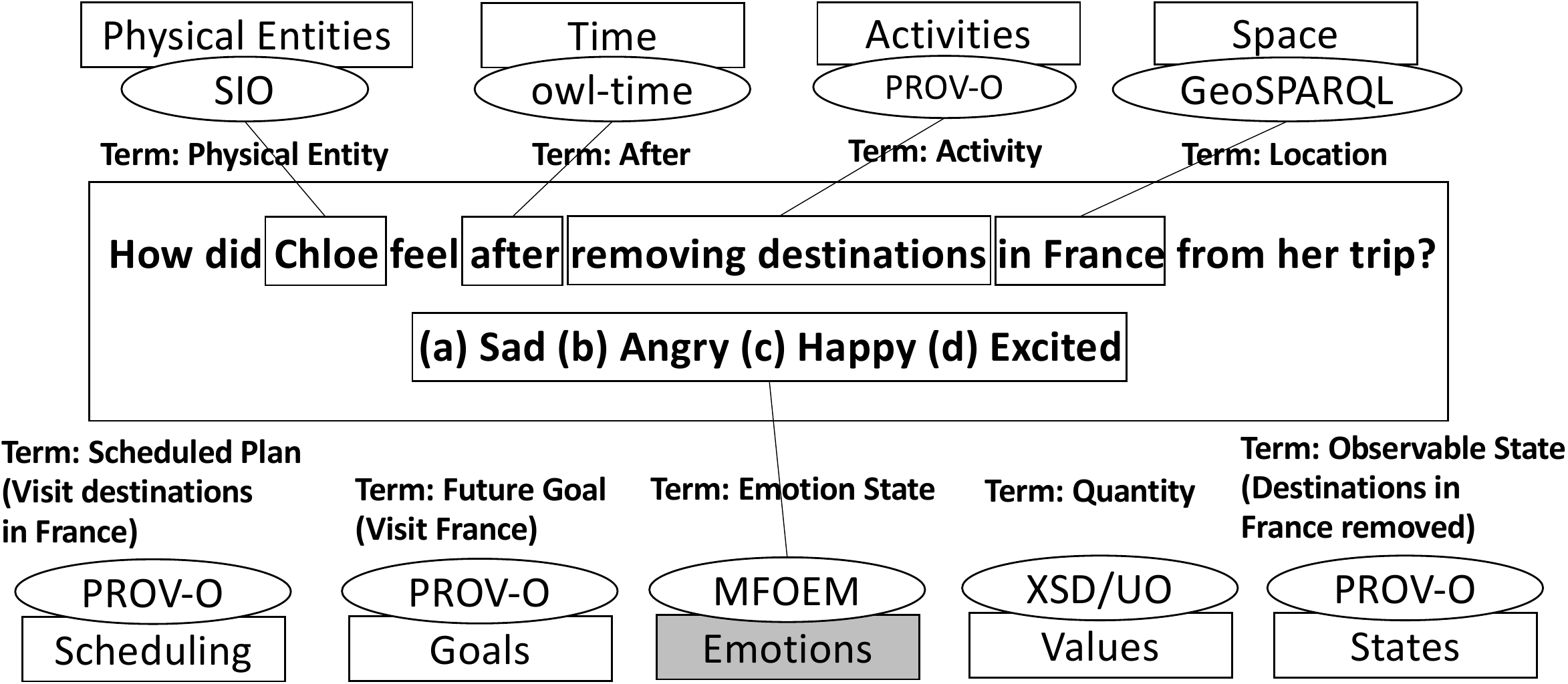}
\caption{\textbf{The Machine Common Sense taxonomy used for TG-CSR QA construction}. The 9 categories are represented in boxes with the associated existing ontologies and vocabularies in ellipses. For each category, the identified elements in the question are linked to the respective formalization. The corresponding term from the Gordon-Hobbs theory is in bold. For the remaining categories, although terms are not explicitly in the text, they can be inferred and are also represented in bold. Because all candidate answers are derived from terms in the ``Emotions'' category, this specific question is categorized as ``Emotions'' in the benchmark, denoted by the gray box.}
\label{fig:taxonomy}
\end{figure}

The Gordon-Hobbs theory defines a representational area of common sense as a cluster of related terms (concepts). Collectively, the 48 representational areas or categories presumably cover all of the identified aspects of human commonsense thinking. Some categories have also been explored in knowledge representation and Semantic Web research. The most prominent example is the ``Time'' category, which has been largely implemented in the \emph{owl-time}~\cite{Little:20:TOO} ontology. Other categories, however, have only been partially formalized in existing ontologies and vocabularies. Figure \ref{fig:taxonomy} denotes the ontologies and vocabularies that we have identified as overlapping with each of the nine selected categories underlying TG-CSR.  We also provide an example of how they are used in the questions and answer options contained in the benchmark.

The ``Time'' category is extensively correlated with the \emph{owl-time} ontology. Both the Gordon-Hobbs theory of time and owl-time base their formalization on Allen's theory of action and time~\cite{allenGeneralTheoryAction1984}. General concepts like \textit{duration}, \textit{end time}, and \textit{start time} are defined in \emph{owl-time}, as well as concepts expressing relationships between time entities like \textit{before}.

Although one could intuitively assume the ``Space'' category to be similar to ``Time'' in relation to its pervasiveness, the most noted representation of ``Space'' is associated with geography. GeoSPARQL~\cite{perryOGCGeoSPARQLAGeographic2012} implements several topological relations, including concepts such as \textit{intersects}, \textit{within}, and \textit{distance}. Some of these predicates map into concepts of the Gordon-Hobbs theory of ``Space''.

For the ``Emotions'' category, Gordon-Hobbs derived many of the concepts from the work of Ortony et al.~\cite{ortony1990} on basic emotions. An existing ``Emotion'' ontology (MFOEM)~\cite{hastings2011} associates many of these basic emotion types with a wide range of other emotions.

The ``Physical Entities'' category is concerned with physical objects in the world, including their shapes and composing parts. The \emph{Semanticscience Integrated Ontology (SIO)}~\cite{dumontierSemanticscienceIntegratedOntology2014a} is a well-known established model that was initially developed for supporting biomedical research. It provides an interesting formalization for material entities that includes concepts like \textit{has part}, \textit{is part of}, and \textit{surrounds}.

The ``Values'' category denotes the use of quantities to measure things, whether quantitative or qualitative. It includes concepts like \textit{quantity}, \textit{value}, and \textit{range}. The XML Schema Definition (XSD)~\cite{Malhotra:12:WXS} supports the usage of data types inside XML documents, including ontology definitions. More than that, XSD formalizes some relationships between quantities, including \textit{maximum}, and \textit{minimum}. The \emph{Units Ontology} (UO)~\cite{gkoutosUnitsOntologyTool2012} provides several units of measurement that can be used to characterize values and quantities.

The remaining categories of ``Goals'', ``Scheduling'', ``States'', and ``Activities'' are closely related to ``Events''. Concepts in these categories include activities, their agents, objectives, and their organization. The PROV Ontology~\cite{McGuinness:13:PTP} provides a formalization that can be used to model activities and associated agents. It also contains predicates that allow the explicit representation of precedence between activities.

\subsection{Terms}

Gordon-Hobbs acknowledge that ``concepts in commonsense knowledge cannot be defined precisely with necessary and sufficient conditions''. In their theory, they identify a list of terms that express concepts associated with each representational area and recommend that these terms should also be explicitly formalized. For instance, in the ``Time'' area, one term is \textit{before}, and it is defined as ``A moment with a lesser position in an ordered set of moments.'' They also formalize \textit{before} as a predicate, e.g., \textit{(before t1 t2)} where some time \textit{t1} is before some time \textit{t2}.

In our work, the formalism presented above was framed as a \emph{taxonomy} from which we used the categories, associated terms, and derived concepts to support the construction of the questions and to restrict the set of answers that could serve as candidates. During benchmark construction, benchmark developers were asked to think about questions where the answers would match concepts in a specific category. The answers could be direct predicates as specified in the taxonomy, or more often, they could be viewed as an instantiation of a concept. For developing the questions, developers were also encouraged to use concepts from multiple categories.

As an example, one of the questions created for the dataset asks: ``How did Chloe feel after removing destinations in France from her trip?'' A plausible answer is ``sad'' which is a defined emotion concept with a defined predicate \textit{(sad p)} that implies that some person is sad. This question is placed in the ``Emotions'' category along with similar questions that can be answered by the answer options associated with various emotion predicates. Figure \ref{fig:taggedquestions} depicts two more examples. The top one shows a question from the ``Time'' category with instantiated time concepts as candidate answers. The question itself contains time concepts used directly, as well as an instantiated activity concept. The bottom one is a question from the ``Activities'' category with instantiated concepts as answers. A direct time concept is used in the question itself.


\section{Question-Answer (QA) Construction}

Given the taxonomy discussed in the previous section, as well as a theme and context, we need to construct questions and candidate answer-sets for each of the nine categories in the taxonomy. Before describing the construction methodology, we note some important desiderata that need to be fulfilled by the QA portion of the benchmark:



\begin{enumerate}
    \item The QA instances must be designed in terms of assessing performance on the major Gordon-Hobbs categories like time and space. 
    \item The files containing the QA instances must have a clear machine-readable structure and format, while still being amenable to human purview. More generally, the files should be easy to work with, with clear documentation, and released on publicly maintained competition platforms. 
    \item The QA instances must be associated with clear annotation guidelines to generate reliable ground truths. Furthermore, the ground truth itself, thus generated, must not only be shown to be stable but also exhibit high human performance (to validate the claim that the benchmark is indeed a commonsense benchmark).
    \item Ideally, the QA instances should be amenable to assessments by both generative and discriminative language representation models (LRMs)~\cite{gpt3,devlin_bert_2019}, although the former is expected to take more manual effort, unless metrics inspired by communities such as machine translation (e.g., BLEU~\cite{bleu}) are used, with an understanding of their limitations for evaluating potentially open-ended answers. 
    \item Given that there are already so many QA benchmarks with `training' and `development' sets openly available, and the \emph{general} nature of commonsense reasoning, the proposed benchmark should be few-shot (or even zero-shot, eventually). In other words, the benchmark must be so designed that good performance from LRMs or other statistical models cannot be explained through careful benchmark-specific fine-tuning alone. In effect, this forces the model to rely on more fundamental aspects of commonsense, including commonsense theory and semantics, to truly perform well on the QA instances. 
\end{enumerate}

\subsection{Methodology}\label{sec:bcon}


\begin{figure}
\centering
\includegraphics[width=\linewidth]{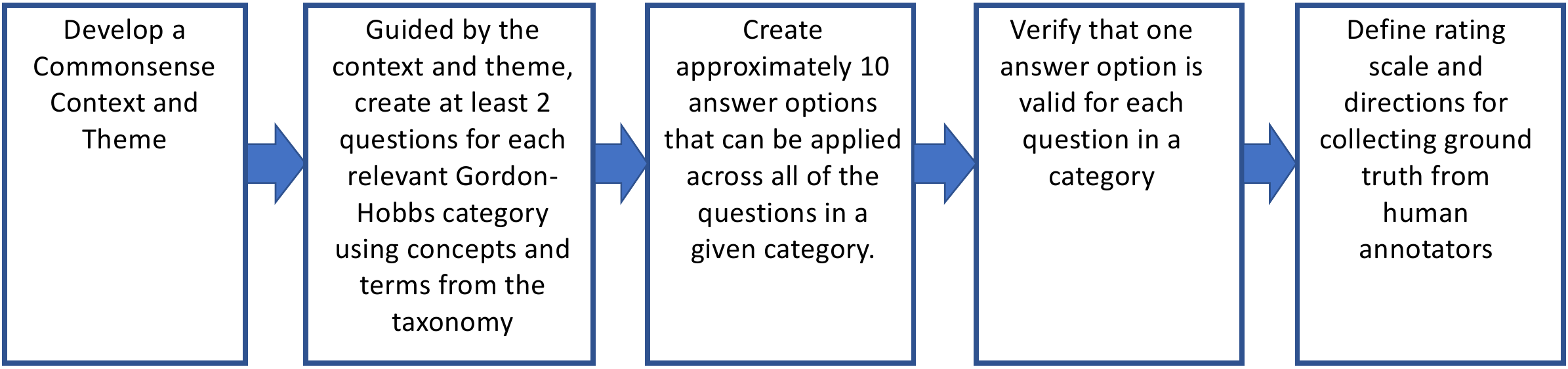}
\caption{A workflow illustrating the key steps of our QA and ground-truth construction methodology.}
\label{fig:processDiagram}
\end{figure}

Figure \ref{fig:processDiagram} summarizes the process that was used to create the datasets and associated ground truth for the first phased release of the benchmark. The guiding motivation behind our QA construction methodology was to emulate complex human interactions in real life, where multiple decisions involving common sense happen in a specific setting, often in a multi-hop manner. In support of this, we developed four contexts and themes. For each context/theme pair, several questions were created that \emph{require} the use of the respective context/theme to be answered correctly. This makes the problem more realistic, but also potentially challenging, for purely statistical models that do not rely on context at all but treat each question as a standalone input. 

\begin{figure}
\centering
\includegraphics[width=\linewidth]{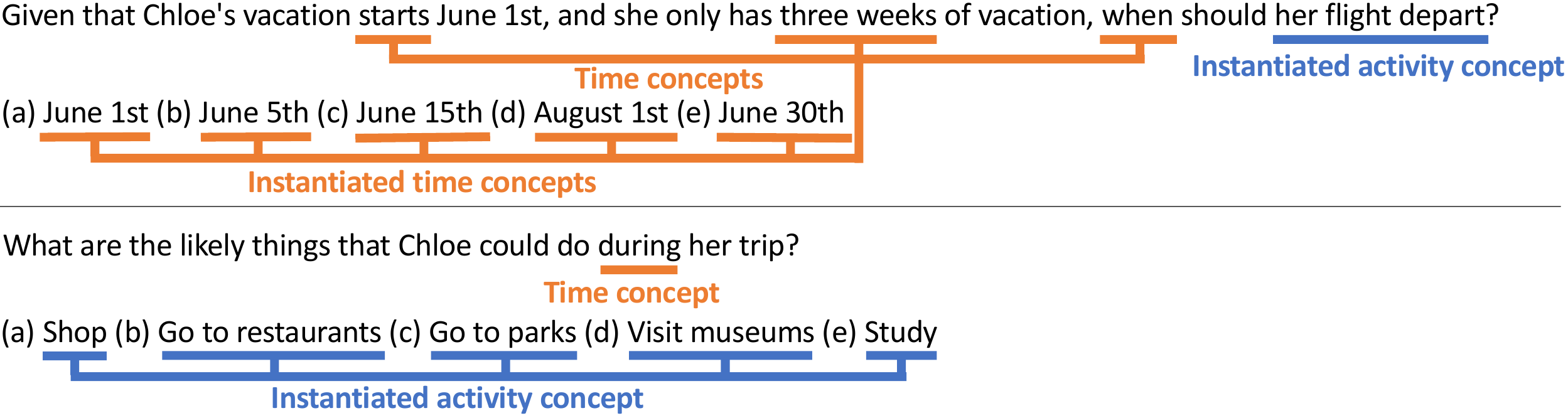}
\caption{Example questions from the ``vacationing abroad'' dataset, with annotated examples of the usage of taxonomy.}
\label{fig:taggedquestions}
\end{figure}

Using the taxonomy introduced earlier, benchmark developers created questions and candidate answers, both ``correct'' and ``incorrect''. Category membership was determined by the answer, and the answer option was influenced by concepts that were incorporated into the taxonomy. As an example, Figure \ref{fig:taggedquestions} contains two questions. In the first question, the answer options are all examples of a potential start time, but not all of these options can satisfy the duration constraints (e.g., three weeks) specified in the question. However, because all answer options are derived from concepts in the ``Time'' category, the question is categorized as ``Time''. In the second question, all answer options are examples of a possible ``activity” that could occur in some time frame, expressed with the term ``during" in the question. Following suit, the question is categorized as ``Activities'' because all answer options were created to agree with concepts in the ``Activities'' category.


In contrast with existing benchmarks, the questions were created as multiple-set, rather than multiple-choice, since there could be more than one correct answer per question. This also makes the benchmark more realistic. Initially, each question had its own set of answer options. Although there could be multiple ``correct'' answers, for any given question, at least one of the answers in the answer option list had to be a ``correct'' answer. To scale the benchmark, all the answers from all of the questions in a given category were combined into a global per-category set. During this process, duplicate answers were removed and, in a few specific cases, the questions or the answers were rewritten. To support benchmark data analysis, a question ID was generated for each question and an answer ID was generated for each answer option.


All benchmark data (questions, per-category candidate answers-set, contexts, themes), and metadata (IDs, category annotations) were constructed in spreadsheets to support benchmark development and annotation. Then, to facilitate LRM experimentation and to make the dataset amenable to a competition-style leaderboard, all benchmark data was converted into several structured (using Javascript Object Notation or JSON) files that could be ingested and processed by a program relatively easily.

\subsection{Ground Truth Development}

 Following initial QA construction, as discussed earlier, the ground truth for the benchmark was derived by averaging the annotations of a \emph{group} of human annotators. By using averaging, rather than annotations from only one individual, we were able to verify high agreement (correlation between annotators was greater than 0.9) and to obtain a more statistically rigorous annotation per QA instance. A more technical description of the averaging procedure is provided at the start of Section \ref{sec:beval}.

Additionally, to prevent possible bias that could influence the annotations, the category names of the questions were obfuscated in the distributed annotation spreadsheets. In the spreadsheets used by the annotators, a tab per category was provided with all questions and answers in their respective tab. However, each tab label was replaced by a label such as ``QuestionTab1'' instead of (for example) ``Time''. We created and provided a concise set of guidelines to the human annotators. In these directions, the human annotators were asked to read the context and theme before answering the questions in each spreadsheet tab. Annotators were told to evaluate each answer option per question on a Likert scale of 1-4 (4: very good fit; 3: good fit; 2: not sure, and 1: bad fit). Annotators could optionally provide comments about uncertainty in answering the question. We will take these comments into account when refining and constructing future releases of the benchmark.

\section{Benchmark Release and Statistics}

\subsection{Benchmark Release}

To make TG-CSR public-facing, we provide an overview on a website, along with data access instructions, and a link to the competition leaderboard. On this leaderboard platform, called CodaLab, the first phase of TG-CSR (with the ``vacationing abroad" context) was recently released and has already been attracting submissions.

The license specification of the resource, and other corresponding links and details (including the website and competition link), are provided in Table \ref{Table: availability}. To access the TG-CSR dataset, users are required to create a free CodaLab account and register for the competition. Once the registration is approved, users are able to log in to participate in the competition by downloading the \emph{Starting kit} and \emph{Public Data}. The \emph{Starting kit} contains a detailed description of the dataset to help users understand the file structure and formats. It also contains starter code to validate the file format that can be accepted by the leaderboard as submission. CodaLab automatically evaluates and scores the submissions, allowing users to obtain performance measures in real-time.

The \emph{Public Data} contains the first of the four datasets (with the ``vacationing abroad" context), including training, development, and test sets. Since TG-CSR is initially designed to be a few-shot problem, we provide labels for the training and development sets, withholding labels for the test set. Users have to submit their predictions for this test set. As discussed earlier, the three subsequent phases of TG-CSR will get progressively more difficult, with the last phase anticipated to be a zero-shot problem\footnote{That is, we do not plan to provide training and development sets in the final phase. }. The three subsequent phases will use the commonsense contexts\footnote{The contexts, themes and QA instances for these three phases have already been constructed but may be refined in minor ways depending on feedback received for the first phase, which is now frozen. } ``bad weather'', ``camping vacation'', and ``dental cleaning'', and are slated for release within 2022. Announcements of these releases will also be posted in a timely manner on the TG-CSR website in Table \ref{Table: availability}.

\begin{table}[]
\centering
\begin{tabular}{p{4.2cm}|p{7.4cm}}
\hline
Benchmark website & \url{https://usc-isi-i2.github.io/TGCSR/}                \\ \hline
\begin{tabular}[c]{@{}l@{}}Canonical citation \\ associated with the resource\end{tabular}                                                     & \url{https://doi.org/10.48550/arXiv.2203.12184}          \\ \hline
Licence specification                                                              & CC BY 4.0                                          \\ \hline
Competition and public data                                                                & \url{https://codalab.lisn.upsaclay.fr/competitions/3080} \\ \hline
\end{tabular}
\caption{Benchmark access and license details.}
\label{Table: availability}
\end{table}


\subsection{Benchmark Statistics}

For the first phased release (i.e., the ``vacationing abroad'' context), there are a total of 331 Q/A pairs in the dataset. The evaluated models yield Yes/No predictions for each Q/A pair, with Yes (encoded using 1) indicating that the provided answer is a good fit for the corresponding question, and No (encoded using 0) indicating a bad fit. The training set, development set, and test set include 81, 77, and 173 such pairs, respectively.  Only the test set does not currently include publicly available labels on the competition leaderboard. As noted earlier, future releases will draw on other contexts and themes, and eventually evaluate zero-shot CSR. Their overall sizes will be similar, but the test sets will be larger since we will be progressively `reducing' the labeled training and development sets in size to make the task more progressively difficult in each phase.  

\section{Benchmark Evaluation}\label{sec:beval}

We use the F1-score for reporting performance on the leaderboard. To evaluate human performance, we used a strategy inspired by leave-one-out cross-validation. Given $k$ human annotators who had each independently graded an answer-question pair on a scale of 1-4, as described earlier, we first `collapsed' 3 and 4 into the `yes' label, and 1 and 2 into the `no' label. This is standard practice when using a finer-grained annotation scheme than the actual labeling scale needed (quaternary vs. binary, in this exercise). Next, we use the labels of each annotator in turn as the \emph{ground-truth}, and take the mode of the remaining $k-1$ annotators as the \emph{human prediction}. We compute an F1-score for this prediction. Repeating the exercise for each annotator ultimately yields $k$ F1-score estimates, the average of which is used as the human performance. In the ``vacationing abroad'' context that is currently released, the human performance (using the procedure above) was found to be 80.5\% for the overall dataset (training, testing, and development), and 79.9\% for only the test set, illustrating high consistency. In contrast, a random baseline, which is computed based on the number of questions in the test set, was found to only achieve 35\% F1-Score. We will similarly compute human performance before releasing the zero-shot contexts in the future. 

\begin{table} \small%
\begin{tabular}{p{2.1cm}p{2cm}p{5.5 cm}p{1cm}p{1.5cm}}
\hline
Model & \#Parameters & Pretraining Data & & Score \\
\midrule
Random &  &  &  & 0.35\\
\midrule
T0-small & 3B  &  \textit{Multiple-Choice QA, Extractive QA}     & &  0.08 \\
         &     &  \textit{closed-Book QA, Structure-To-Text} & & \\
         &     &  \textit{Sentiment Analysis, Summarization} & & \\
         &     &  \textit{Topic Classification, and Paraphrase Identification benchmarks} &  \\
\midrule
T0               & 11B & \textit{(same as T0-small) \^}  & & 0.15 \\
\midrule
T0+              & 11B & \textit{same as T0 \^  plus additional datasets from GPT-3's~\cite{NEURIPS2020_1457c0d6} evaluation suite} & & 0.0 \\
                
\midrule
T0++ & 11B &   \textit{Same as T0+ \^  plus additional datasets from the SuperGLUE benchmark}~\cite{NEURIPS2019_4496bf24} & & 0.603\\
\midrule
Human & & & & 0.799  \\
\midrule
\end{tabular}
\caption{Baseline results of the T0* language representation models on our benchmark. As illustrated, the largest model with the most pretraining (T0++) data achieves the highest score, which is still far from human performance, showcasing the utility of TG-CSR. Models are downloadable from the \href{https://huggingface.co/bigscience/T0pp}{Hugging Face repository}.}
\label{tbl:baselines}
\end{table}

We also benchmarked the dataset using the family of T* language representation models based on a recent paper~\cite{sanh_multitask_2022}. 
These models are prompt-based encoder-decoder models fine-tuned on a large collection of datasets (also described in \cite{sanh_multitask_2022}) in a multi-task manner.  The T0 models have demonstrated solid zero-shot performance on various natural language processing tasks, including popular CSR benchmarks, making them a good candidate for the evaluation of our benchmark.
We experiment with two sizes of these models, T0-3B, and T0-11B, which encode contain 3 billion and 11 billion parameters. Moreover, to obtain the best possible performance on the benchmark, we use different versions of the 11B parameter models and report the best results in Table \ref{tbl:baselines}.

The highest score of 60.3\% F1-Score was achieved by the model T0++. While promising, this result also suggests that there is a considerable distance between a reasonable LRM's performance and human performance (79.9\%). 
Consistent with previous studies~\cite{JMLR:v21:20-074,sanh_multitask_2022,NEURIPS2019_4496bf24}, the \emph{largest model} with the \emph{most pretraining data} (T0++) achieves the highest performance on the benchmark.  Further analysis of the models' predictions on TG-CSR suggests that they are biased toward classifying most of the question-answer pairs as ``No'', which explains the low F1 scores. It also validates the use of the F1-score, rather than simple accuracy (fraction of correct answers).
This observation further emphasizes the importance of creating more diverse datasets and benchmarks as resources both for training and evaluation purposes for the models. It also provides evidence supporting our earlier claim that, by virtue of being more theoretically and semantically grounded, TG-CSR is more challenging for purely statistical models (at least at the present moment). 

\section{Discussion}


During the development of question and associated answer options, we observed some differences in opinion by both the dataset developers and the annotators on what is (or is not) considered to be commonsense. For example, while creating questions and answer options about how to pack a suitcase, we encountered different preferences for items to pack and for the packing order.  We assume that the source of the difference is likely associated with differences in the age, sex, and cultural background of the dataset developers.   Similarly, in another question about bringing food for a picnic, we observed differences between what types of food and drinks were considered good examples of commonsense choices, e.g., bringing wine was considered questionable by some\footnote{In the United States, and many other countries, consumption of alcohol in open public spaces like parks is usually illegal.}.  Where we could not reach any agreement on a particular QA instance, we removed the QA instance from the dataset prior to release.  However, our observations suggest, perhaps unsurprisingly, that knowledge of a personal or cultural type can intersect in non-trivial ways with commonsense knowledge, and the two are not always easy to distinguish theoretically.  In the picnic example, for instance, the commonsense knowledge appears to exist at a higher level of abstraction.  More experiments are needed to understand these observations in depth. If commonsense knowledge indeed exists at a higher level of abstraction, it might suggest that, rather than provide physical entity values as answer options for these questions in the dataset, answer choices that are more general or abstract could be more appropriate: e.g., a drink versus a soda, a fruit versus strawberries, and so on.

In a previous experiment~\cite{santosExperimentalStudyMeasuring2021}, annotators reported difficulty in classifying sentences about human emotions. However, in the development of the ``vacationing abroad'' dataset, we noticed that it was quite easy to create potential questions and answers for the ``Emotions'' category.  We attribute this to the fact that we limited the number of candidate choices for the ``Emotions'' questions to ten. This in turn indicates the importance of \emph{constraints} when designing and evaluating questions that might elicit a broad range of responses (at least in terms of vocabulary alone), and the difficulty in potentially evaluating answers in a generative (i.e., open-ended, where answer choices are not provided at all) setting using metrics such as BLEU~\cite{bleu}.

In addition, our development efforts identified several challenges associated with using the Gordon-Hobbs theory and the larger taxonomy to construct a commonsense dataset. For example, there was considerable internal discussion among the benchmark developers regarding what category a question should be placed into. This was particularly true for questions associated with time, activities, and scheduling since they all have temporal aspects associated with them. For example, an answer to a question about when to schedule some activity could be a time value, e.g., \textit{10:00 AM}, or it could be some named activity, e.g., \textit{after eating breakfast}. To resolve these issues, we concluded that the answer value type, e.g., \textit{June 4}, of a question, determines its categorical membership e.g., \textit{time}. We did this to enforce consistency in QA construction, rather than make normative claims about the Gordon-Hobbs theory itself. 

\subsection{Generative Language Models}
\label{subsection-generative}

\begin{figure}
\centering
\includegraphics[width=\linewidth]{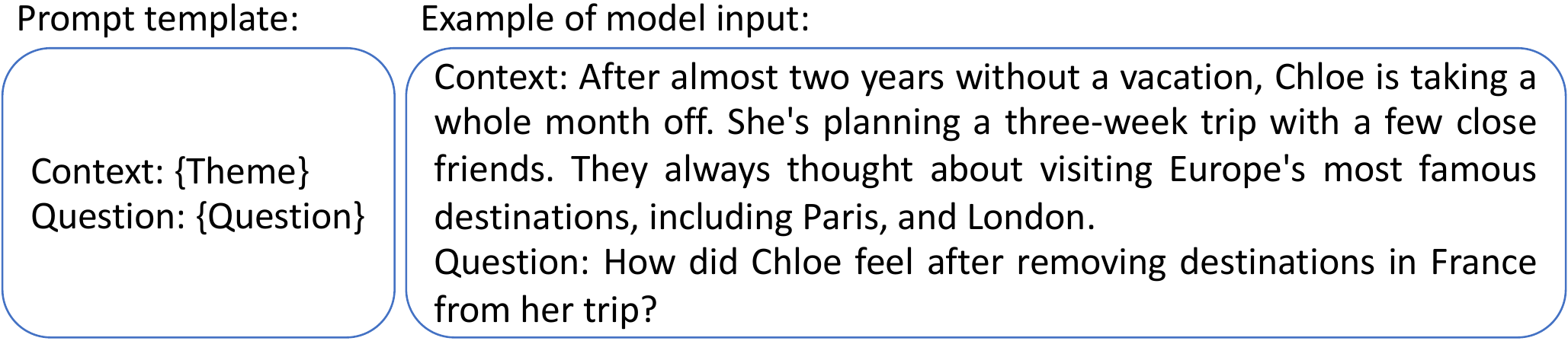}
\caption{Example of prompt template for generative evaluation of the benchmark using language models.}
\label{fig:generative_prompt}
\end{figure}


Although the performance of the language models in Table \ref{tbl:baselines} is well below human performance, we had also observed in subsequent analysis that language models generally returned ``No'' for a QA instance, even when human annotators considered it to be a ``Yes''. This begs the question of whether, given the opportunity, a \emph{generative} language model might return a good answer for a question when it is \emph{not} prompted with pre-determined answer choices. 

To evaluate the generative performance of a language model on TG-CSR, we again use the T0++ model~\cite{sanh_multitask_2022} but in a generative manner. The prompt template, and example of model input, is provided in Figure \ref{fig:generative_prompt}. As shown therein, we concatenated the theme and question and conditioned the model output on the prompt template; this is a common practice in the generative use of these models~\cite{sanh_multitask_2022}. 
Since many of the generated answers did not match any of the answer options provided, we performed a manual review of the results.  To measure the accuracy of a generative answer, we first look for exact or almost-exact matches between the generative answer and any answer options associated with a question.  If there is a match, we then evaluate it by checking the related ground truth values.  In the experiment, six generative answers closely matched a ``Yes'' (or good-fit) ground truth answer option in the following categories: Goals, Physical Entities, and World States; while one generative answer exactly matched one of the ground truth ``No'' answer options in the ``Values and Quantities'' category.

Further review of the answers that were not close matches with any of the answer options (whether ``Yes'' or ``No'') showed that many of the generated answers were plausible fits, although there were also a few implausible fits. For example, the model generated an answer such as ``upset'' for the example question in Figure \ref{fig:overview}. This is not a provided answer choice, but it comports well with the other answer choices, such as ``frustrated'' and ``angry''.  While preliminary, this experiment suggests that, in the future, generative language models could potentially be used to aid in QA construction by supplementing human answers. 



\section{Conclusion and Future Work}

In this paper, we described TG-CSR, a theoretically-grounded benchmark that enables semantic evaluation of machine commonsense reasoning (CSR) abilities. TG-CSR is, to our knowledge, the first CSR benchmark to be developed that is grounded in a formal theory of commonsense, based on the influential work of Gordon and Hobbs. In addition to the actual resource, which is now hosted on a public competition-style leaderboard and will involve phased, and progressively more difficult, releases, we described how existing ontologies and vocabularies can formally represent foundational commonsense categories. We also discussed how this unified commonsense representation can be used to support the creation of theoretically-grounded QA benchmarks. Specifically, concepts in the identified ontologies were combined into a simple taxonomy that guided the creation of questions and answers that compose TG-CSR.


In the future, besides releasing three more progressively difficult phases of TG-CSR, we plan to continue developing and deepening our initial taxonomy to cover \emph{additional} aspects of the Gordon-Hobbs theory, potentially for supporting new benchmarks developed around specific categories, such as planning and organization. Existing CSR benchmarks are not believed to be covering these categories, offering a promising avenue for future benchmark development. We will also continue to refine our ground-truth annotation procedures and add more formal semantics to the existing TG-CSR questions and answers.


\subsubsection{Acknowledgements.}

This work was funded under the DARPA Machine Common Sense (MCS) program under award number N660011924033.

\bibliography{typeinst}
\bibliographystyle{abbrv}
\end{document}